\pdfoutput=1
\documentclass[10pt,twocolumn,letterpaper]{article}

 \usepackage[pagenumbers]{cvpr} 

\usepackage[dvipsnames]{xcolor}

\definecolor{cvprblue}{rgb}{0.21,0.49,0.74}
\usepackage[pagebackref,breaklinks,colorlinks,citecolor=cvprblue]{hyperref}

\def\paperID{*****} 
\def\confName{CVPR}
\def\confYear{2024}

\title{HSEmotion Team at the 6th ABAW Competition: Facial Expressions, Valence-Arousal and Emotion Intensity Prediction} 

\author{Andrey V. Savchenko\textsuperscript{1,2}\\
\textsuperscript{1}Sber AI Lab\\
Moscow, Russia\\
\textsuperscript{2}HSE University\\
Laboratory of Algorithms and Technologies for Network Analysis, Nizhny Novgorod, Russia\\
{\tt\small avsavchenko@hse.ru}
}

\begin{document}
\maketitle
\begin{abstract}
 This article presents our results for the sixth Affective Behavior Analysis in-the-wild (ABAW) competition. To improve the trustworthiness of facial analysis, we study the possibility of using pre-trained deep models that extract reliable emotional features without the need to fine-tune the neural networks for a downstream task. In particular, we introduce several lightweight models based on MobileViT, MobileFaceNet, EfficientNet, and DDAMFN architectures trained in multi-task scenarios to recognize facial expressions, valence, and arousal on static photos. These neural networks extract frame-level features fed into a simple classifier, e.g., linear feed-forward neural network, to predict emotion intensity, compound expressions, action units, facial expressions, and valence/arousal. Experimental results for five tasks from the sixth ABAW challenge demonstrate that our approach lets us significantly improve quality metrics on validation sets compared to existing non-ensemble techniques.
\end{abstract}

\section{Introduction}
\label{sec:intro}

The ability to accurately analyze human emotions is crucial for developing human-centered technologies~\cite{erol2019toward,chowdary2023deep,zhang2023dual}. Contemporary research on affective behavior analysis in unconstrained environments reached a high level of maturity~\cite{zhang2023abaw5}. It is incredibly challenging for the in-the-wild domain, where conventional approaches often struggle due to variations in lighting, pose, and expression intensity~\cite{kollias2019deep,kollias2021affect}. One of the well-known benchmarks for evaluating progress in this field is the sequence of Affective Behavior Analysis in-the-wild (ABAW) competitions~\cite{zafeiriou2017aff,kollias2020analysing,kollias2021analysing,kollias2022abaw,kollias2023abaw2,kollias20246th}. 

The main tasks in the previous editions of ABAW are frame-level video-based recognition of facial expressions (EXPR), valence/arousal (VA), and detection of action units (AU)~\cite{kollias2023abaw}. A simple fusion of Resnet50, Regnet, and EfficientNet backbones significantly improves the baseline~\cite{wang2023facial}. Very high quality was obtained by several teams~\cite{zhou2023continuous,savchenko2022cvprw,savchenko2023cvprw} that used EfficientNet-based model~\cite{savchenko2022classifying} pre-trained on the AffectNet dataset~\cite{mollahosseini2017affectnet}. An affine module was proposed~\cite{zhang2023facial} to align the features to the same dimension in the audio-visual transformer. The audio-visual fusion model that leverages transformer and TCN (Temporal Convolutional Networks) leads to top-3 results in the EXPR competition~\cite{zhou2023continuous}. One of the best results was obtained by the audio-visual transformer-based ensemble of several models, including the facial processing with the fine-tuned masked autoencoder (MAE)~\cite{zhang2023abaw5}. The fusion of audio and video modalities using channel attention is examined in~\cite{zhangSu2023abaw5}. High accuracy of AU detection was achieved by spatio-temporal representation learning~\cite{wang2023abaw5} with MAE and temporal graph embeddings~\cite{makarov2022temporal}.

Another task introduced in the previous ABAW challenge is the Emotional Mimicry Intensity (EMI) estimation~\cite{kollias2023abaw2}, which is the multi-task regression problem that should be solved for the entire video from the Hume-Reaction dataset. The third place was obtained by the above-mentioned MAE~\cite{zhang2023abaw5}. In the paper of the second place winnder~\cite{Yu_2023_CVPR}, the spatial attention mechanism and the Mel-Frequency Cepstral Coefficients were used to extract visual and acoustic features, respectively. At the same time, the temporal dynamics were modeled using TCN and a transformer encoder. Finally, the top performance was achieved by studying, analyzing, and combining diverse models and tools
to extract multimodal features~\cite{Li_2023_CVPR}. However, the organizers of the current ABAW challenge~\cite{kollias20246th} decided to significantly reduce the training and validation sets in this task to make it much more complicated.

However, one major challenge in facial expression recognition is the need for extensive fine-tuning of deep neural networks for specific tasks. For example, the winner of the previous challenge led to excellent results by three times fine-tuning the MAE encoder on the image frames from the Aff-wild2 dataset for AU, EXPR, and VA tasks individually~\cite{zhang2023abaw5}. This procedure leads to obtaining state-of-the-art models for particular datasets but can be computationally expensive and limit the generalizability of models. Thus, the emotion analysis in-the-wild primary goal is to construct single models~\cite{kollias2019face} that are fair, explainable, trustworthy, and privacy-conscious, achieving high performance while enhancing generalization in real-world scenarios. To encourage the research of this final goal, the sixth ABAW competition introduces the new task for unsupervised Compound Expression (CE) recognition~\cite{kollias2023multi} without the labeled training set, so the high-accurate pre-trained models should be utilized in the domain adaptation/self-supervised/zero-shot learning techniques.

To achieve the above-mentioned objectives, we introduce a novel methodology centered around lightweight models derived from architectures such as MobileViT (Mobile Vision Transformer)~\cite{mehta2021mobilevit}, MobileFaceNet~\cite{chen2018mobilefacenets}, EfficientNet~\cite{tan2019efficientnet}, and DDAMFN (Dual-Direction Attention Mixed Feature Network)~\cite{zhang2023dual}. These models are trained in a multi-task scenario, enabling them to discern facial expressions, valence, and arousal from static photographs. By extracting frame-level features, we feed these neural networks into a straightforward classifier, such as a linear feed-forward neural network. Our approach facilitates the prediction of emotion intensity, compound expressions, action units, facial expressions, and valence/arousal, thereby offering a holistic analysis of affective behavior. Our method aims to extract robust emotional features directly applicable to various tasks within the ABAW challenge.

\section{Methodology}
\label{sec:proposed}

\subsection{Multi-Task Learning of Emotional Descriptors}
\label{subsec:training}
In this paper, we use the following approach to train neural networks that extract emotional embeddings~\cite{savchenko2022classifying,zhang2023abaw5}. At first, the neural network is pretrained on a face recognition task. In particular, we use the VGGFace2 dataset~\cite{cao2018vggface2} with 3,067,564 photos of 9131 persons while the validation set contains 243,722 remaining images. The conventional softmax cross-entropy was optimized by Adam and sharpness-aware minimization~\cite{liu2022towards} during 10 epochs. In contrast to studies of face identification~\cite{an2022killing}, we use the cropped facial regions without any margins and alignment to concentrate on the main part of the face. 

Next, we fine-tune the model to recognize emotions on static images from the AffectNet dataset~\cite{mollahosseini2017affectnet} with $C_{(Expr)}=8$ emotional classes corresponding to Anger, Contempt, Disgust, Fear, Happiness, Neutral, Sadness and Surprise, and values of Valence/Arousal from the range between -1 and 1. The official training set contains 287651 manually labeled photos, while the validation set consists of 4000 images (500 per class). We leverage the multi-task (MT) learning~\cite{kollias2019expression,kollias2021distribution} and minimize the CCC (Concordance Correlation Coefficient) for valence/arousal and weighted cross-entropy for facial expressions to mitigate the class imbalance in the training set~\cite{savchenko2022mt}:
\begin{multline}
\label{eq:1}
 L\left(X,y_{(Expr)},y_V,y_A\right) =1- \\ -\log \left(softmax(z_{y_{(Expr)}}) \cdot \underset{y \in \{1,...,C_{(Expr)}\}} \max N_y / N_{y_{(Expr)}}\right)-\\- 0.5\left(CCC(z_V,y_V)+CCC(z_A,y_A)\right),
\end{multline}
where $X$ is the facial image, $y_V$, $y_A$ and $y_{expr} \in \{1,..., C_{(Expr)}\}$ are its valence, arousal and facial expression label, respectively, $N_y$ is the total number of training examples of the $y$-th expression, $z$ is the logits at the output of last fully connected layer.

Due to privacy issues, it is desirable to implement facial analysis on the edge/mobile device~\cite{chowdary2023deep,demochkina2021mobileemotiface,kharchevnikova2018neural}). Hence, we decided to use several well-known lightweight architectures of neural networks, such as MobileViT~\cite{mehta2021mobilevit}, MobileFaceNet~\cite{chen2018mobilefacenets}, and DDAMFN~\cite{zhang2023dual}. The resulting models are called MT-EmotiMobileViT, MT-EmotiMobileFaceNet, and MT-DDAMFN. We made the weights publicly available in the repository with our previous EmotiEffNet models~\cite{savchenko2022classifying,savchenko2022mt}.

\subsection{Video-based VA Estimation, EXPR Recognition, and AU Detection}
\label{subsec:video_model}
The main idea of this paper is to study the usage of pre-trained deep neural networks without fine-tuning them on every downstream task. Though such an approach cannot produce state-of-the-art results for a concrete dataset, it reflects the practically essential requirement for an emotion analysis model that can be used in unconstrained environments. Hence, similar feed-forward neural networks are used for all three tasks of the ABAW challenge with the Aff-Wild2 dataset~\cite{kollias2019expression}, namely, VA estimation, facial expression recognition, and AU detection. These models are trained using all labeled frames from the training sets of corresponding contests. 

The input for EXPR and AU tasks is the embeddings $\mathbf{x}$ at the output of the penultimate layer of our neural network, while the valence and arousal are more accurately predicted using the logits $\mathbf{z}$ at the output of the last layer~\cite{savchenko2023cvprw}. Moreover, the linear model without hidden layers is used in the latter case, while the former case is solved by the multi-layer perceptron with one hidden layer and 128 units. In addition to our models, we computed embeddings extracted by the wav2vec 2.0 model from the audio of the corresponding video~\cite{baevski2020wav2vec}. As the number of audio and video frames is not the same, we mapped the features of an acoustic frame to the closest to the video frame.

The Adam optimizer was leveraged in all cases. As our models typically deal with high-resolution facial images, we processed the folder of cropped faces officially released by the organizer of the 6th ABAW competition. For VA estimation, tanh activations were added to the output layer, and CCC loss was optimized for 20 epochs. The training set contains 1,653,930 cropped faces from 356 videos, while validation is performed on 376,332 other images from 76 videos~\cite{kollias20246th}.

For EXPR classification, we use softmax activations and weighted categorical cross-entropy with weights inversely proportional to the number of examples of each class in the training set. The latter contains 585317 labeled frames from cropped faces of 248 videos provided by the organizers, while the testing is performed on 280,532 validation images from 70 videos. The model is trained for 10 epochs.

Finally, sigmoid activations and multi-class weighted binary cross-entropy were utilized for AU detection. The training set consists of 1,356,694 labeled facial frames of 295 videos, and 445,836 images from the remaining 105 videos are included in the validation set.

As the decision is made in a frame-level manner, some noise may be introduced in the outputs of trained feed-forward neural networks. Thus, we smoothed the predictions for each frame in a short window~\cite{savchenko2023cvprw} using a box filter, i.e., the arithmetic mean of predicted scores for all frames in a window. AU units may rapidly change, so a short window with 5 frames (2 before and after the current frame) was utilized. Facial expressions and valence and arousal last for longer periods. Hence, the windows for these tasks contain 50 frames in our experiments.

\subsection{Compound Expression Recognition}
\label{subsec:ce_model}
The new task of the 6th ABAW competition is the frame-wise CE recognition on videos from the C-EXPR database~\cite{kollias2023multi}. It is required to assign each frame of 56 videos into one of 7 classes, namely, Fearfully\_Surprised, Happily\_Surprised, Sadly\_Surprised, Disgustedly\_Surprised, Angrily\_Surprised, Sadly\_Fearful, Sadly\_Angry. It is the most complicated task as there is no labeled validation set to compare different solutions. In order to somehow choose the candidate submissions, we decided to measure the class balance. The authors of the C-EXPR~\cite{kollias2023multi} presented the number of frames for each compound class: 14445, 24915, 10780, 10637, 10535, 10112, and 8878. Hence, we choose the Kullback-Leibler divergence between the frequencies of each class at the output of our models and the frequencies of classes computed from these number of frames.

The faces from each frame were extracted with the RetinaFace model~\cite{deng2020retinaface}. If it cannot detect the face on a particular frame, it feeds into the input of the face detector from the Mediapipe framework~\cite{lugaresi2019mediapipe}. As a result, we obtained 22,641 frames with at least one video. It is possible that several faces are detected for a frame, so the total number of detected faces is equal to 32,329. We analyze all of them with our emotional models. Next, we predict $C_{(Expr)}=8$ AffectNet's basic expressions for every detected face, compute probabilities from logits $\mathbf{z}$, and summarize the probability scores for two classes from the compound expressions. Predictions for several faces inside one frame are aggregated with the arithmetic mean. The final prediction is the compound class label corresponding to the maximal summary score. 

In addition, we decided to use clustering of embeddings extracted from video and audio frames. Simple K-means clustering with 7 clusters is utilized. To choose the label of each cluster, we compute the average scores of compound classes for all frames from each cluster. The scores are computed as described in the previous paragraph. 

\subsection{Emotional Mimicry Intensity Estimation}
\label{subsec:emi_model}
The EMI estimation is a multi-output regression problem with six categories (Admiration, Amusement, Determination, Empathic Pain, Excitement, and Joy. In contrast to previous tasks, one label per whole video is available. Hence, it is necessary to obtain a single descriptor for an entire video, given the facial features of every frame. The official training set contains 8072 videos, while 4588 videos are available for validation.

In this paper, we used simple STAT (statistical) features that have previously shown excellent performance in EmotiW (Emotion recognition in-the-Wild) challenges~\cite{bargal2016emotion,demochkina2021mobileemotiface}. In particular, we compute component-wise mean, standard deviation, minimum, and maximum of logits/embeddings at the output of our model and concatenate them into a single descriptor. The latter is fed into a linear classifier (feed-forward neural network without hidden units) with six outputs and sigmoid activation functions. The weighted Pearson Correlation Coefficient (PCC) $\rho$ loss is minimized for logits $\mathbf{z}$ by the Adam with 100 epochs. If the embeddings of our models or wav2vec 2.0 features are estimated, they are fed into a multi-layer perceptron with one hidden layer and 128 units. The output scores of the trained neural net without any post-processing are directly used as the final predictions. 

\section{Experimental Results}
\label{sec:exper}

\subsection{Facial Emotion Analysis for Static Photos}

\begin{table*}
    \centering
    \begin{tabular}{ccccccc}
        \toprule
        & \multicolumn{2}{c}{Facial expressions} & \multicolumn{2}{c}{Valence} & \multicolumn{2}{c}{Arousal} \\
Model & 8-Acc., \% ($\uparrow$)& 7-Acc., \% ($\uparrow$) & RMSE ($\downarrow$) & CCC ($\uparrow$)& RMSE ($\downarrow$)& CCC ($\uparrow$)\\
        \midrule
AlexNet~\cite{mollahosseini2017affectnet} & 58.0 & -& 0.394 & 0.541 & 0.402 & 0.450\\
SSL inpanting-pl~\cite{pourmirzaei2021using} & 61.72  & -& - & - & - & -\\
Distract Your Attention~\cite{wen2021distract} & 62.09  & 65.69 & - & - & - & -\\
ViT-base + MAE~\cite{li2023emotion} & 62.42 & -& - & - & - & -\\
Static-to-Dynamic~\cite{chen2023static} & 63.06 & 66.42 & - & - & - & -\\
DDAMFN~\cite{zhang2023dual} & 64.25	& 67.03 & - & - & - & -\\
\hline
EmotiEffNet-B0 & 61.32 & 64.57& - & - & - & - \\
MT-EmotiEffNet & 61.93 & 64.97 & 0.434 & 0.594 & 0.387 & 0.549\\ 
MT-EmotiMobileFaceNet & 62.32 & 65.17 & 0.447 & 0.577 & 0.387 & 0.547\\
MT-EmotiMobileViT & 62.50 & 66.46 & 0.423 & 0.599 & 0.371 & 0.565\\
EmotiEffNet-B2 & 63.13  & 66.51 & - & - & - & -\\
MT-DDAMFN & 64.20 & 67.00 & 0.363 & 0.729 & 0.341 & 0.643\\
        \bottomrule
    \end{tabular}
    \caption{Results for the AffectNet validation set (high Accuracy and CCC are better, low RMSE is better)}
    \label{tab:affectnet}
\end{table*}

\begin{table*}
 \centering
 \begin{tabular}{cccccc}
 \toprule
Method & Modality & Is ensemble? & CCC\_V & CCC\_A& $P_{VA}$ \\
 \midrule
 Baseline ResNet-50~\cite{kollias20246th} & Faces & No & 0.24 & 0.20 & 0.22\\
 \hline
 EfficientNet-B0~\cite{savchenko2022cvprw} & Faces & No & 0.449 & 0.535 & 0.492\\
 Resnet50/Regnet/EfficientNet~\cite{wang2023facial} & Faces & Yes &  0.257 & 0.383 & 0.320 \\ 
Channel Attention Network~\cite{zhangSu2023abaw5} & Audio/video & Yes & 0.423 & 0.670 & 0.547 \\
MAE~\cite{zhang2023abaw5} & Audio/video & Yes & 0.476 & 0.644 & 0.560\\
Transformer~\cite{zhang2023facial} & Audio/video & Yes &  0.554 & 0.659 & 0.607 \\ 
TCN~\cite{zhou2023continuous} & Audio/video & Yes & 0.550 & 0.681 & 0.615 
\\
 \hline
 wav2vec 2.0 & Audio & No & -0.011 & 0.244 & 0.116 \\
EmotiEffNet-B2 & Faces & No & 0.423 & 0.498 & 0.464 \\
EmotiEffNet & Faces & No & 0.443 & 0.519 & 0.482 \\
EmotiEffNet, smoothing & Faces & No & 0.490 & 0.596 & 0.543 \\
MT-EmotiEffNet & Faces & No & 0.444 & 0.521 & 0.483 \\
MT-EmotiEffNet, smoothing & Faces & No & 0.490 & 0.604 & 0.547 \\
DDAMFN & Faces & No & 0.438 & 0.523 & 0.481 \\
DDAMFN, smoothing & Faces & No & 0.485 & 0.598 & 0.541 \\
MT-EmotiMobileViT & Faces & No & 0.445 & 0.525 & 0.485 \\
MT-EmotiMobileViT, smoothing & Faces & No & 0.493 & 0.612 & 0.552 \\
MT-EmotiMobileFaceNet & Faces & No & 0.439 & 0.532 & 0.486 \\
MT-EmotiMobileFaceNet, smoothing & Faces & No & 0.483 & 0.610 & 0.547 \\
MT-DDAMFN & Faces & No & 0.468 & 0.537 & 0.502 \\
MT-DDAMFN, smoothing & Faces & No & 0.519 & 0.616 & 0.568 \\
 \bottomrule
 \end{tabular}
 \caption{Valence-Arousal Challenge Results on the Aff-Wild2’s validation set.}
 \label{tab:va}
\end{table*}

\begin{table*}
 \centering
 \begin{tabular}{ccccc}
 \toprule
 Method & Modality & Is ensemble? & F1-score $P_{EXPR}$ & Accuracy \\
 \midrule
 Baseline VGGFACE (MixAugment)~\cite{kollias20246th} & Faces & No & 0.25 & -\\
\hline
EfficientNet-B0~\cite{savchenko2022cvprw} & Faces & No & 0.402 & - \\
Meta-Classifier~\cite{wang2023facial} & Faces & Yes &  0.302 & 0.462 \\ 
TCN~\cite{zhou2023continuous} & Audio/video & Yes& 0.377 & -  \\
Transformer~\cite{zhang2023facial} & Audio/video & Yes & 0.406 & - \\ 
MAE~\cite{zhang2023abaw5} & Audio/video & Yes & 0.495 & - \\ 
\hline
wav2vec 2.0 & Audio & No & 0.291 & 0.410 \\
wav2vec 2.0, smoothing & Audio & No & 0.355 & 0.521 \\
DDAMFN & Faces & No & 0.308 & 0.433 \\
EmotiEffNet-B2 & Faces & No & 0.320 & 0.438 \\
MT-EmotiMobileFaceNet & Faces & No & 0.327 & 0.462 \\
MT-EmotiEffNet & Faces & No & 0.336 & 0.447 \\
MT-DDAMFN & Faces & No & 0.351 & 0.469 \\
MT-EmotiMobileViT & Faces & No & 0.356 & 0.461 \\
EmotiEffNet& Faces & No & 0.384 & 0.495 \\
EmotiEffNet, smoothing & Faces & No & 0.424 & 0.543 \\
wav2vec 2.0+EmotiEffNet& Audio/video & Yes & 0.403 & 0.520 \\
wav2vec 2.0+EmotiEffNet, smoothing & Audio/video & Yes & 0.434 & 0.557 \\
 \bottomrule
 \end{tabular}
 \caption{Expression Challenge Results on the Aff-Wild2’s validation set.}
 \label{tab:expr}
\end{table*}

\begin{table*}
 \centering
 \begin{tabular}{cccc}
 \toprule
 Method & Modality & Is ensemble? & F1-score $P_{AU}$ \\
 \midrule
 Baseline VGGFACE~\cite{kollias20246th} & Faces & No & 0.39\\
\hline
IResnet100~\cite{yu2023local} & Faces & Yes& 0.511\\ 
TCN~\cite{zhou2023continuous} & Audio/video & Yes& 0.517 \\
Transformer~\cite{zhang2023facial} & Audio/video & Yes & 0.530\\ 
Regnet/Video Vision Transformer~\cite{ngoc2023abaw5} & Faces & No & 0.540\\ 
Masked Autoencoder graph representations~\cite{wang2023abaw5} & Faces & Yes& 0.543 \\ 
Masked Autoencoder~\cite{zhang2023abaw5} & Audio/video & Yes& 0.567\\
   Regnet~\cite{wang2023facial} & Faces & Yes& 0.698 \\ 
 \hline
wav2vec 2.0 & Audio & No & 0.313 \\
DDAMFN & Faces & No & 0.500 \\
MT-EmotiMobileFaceNet & Faces & No & 0.512 \\
MT-DDAMFN & Faces & No & 0.519 \\
MT-EmotiEffNet & Faces & No & 0.525 \\
EmotiEffNet & Faces & No & 0.537 \\
EmotiEffNet, smoothing & Faces & No & 0.545 \\
 \bottomrule
 \end{tabular}
 \caption{Action Unit Challenge Results on the Aff-Wild2’s validation set.}
 \label{tab:au}
\end{table*}

In the first experiment, we demonstrate the efficiency of our models for the official validation part of AffectNet~\cite{mollahosseini2017affectnet}. Table~\ref{tab:affectnet} contains the RMSE (Root Mean Square Error) and CCC for predicted valence/arousal and accuracy for facial expression recognition. In the latter case, we compute two metrics traditionally used with this dataset, namely, 8-Acc (Accuracy for all eight classes) and 7-Acc (Accuracy for seven basic categories: Anger, Disgust, Fear, Happiness, Neutral, Sadness and Surprise). Existing papers usually train two different models with 8 and 7 outputs and report the performance of each model separately. However, our primary goal is to study the universality of our models. Hence, we train only one model with eight emotional categories and remove the logits corresponding to Contempt to obtain the outputs for seven basic classes.

As one can see, our models trained with the multi-task loss (\ref{eq:1}) show very high performance. The state-of-the-art DDAMFN~\cite{zhang2023dual} is only slightly more accurate when compared to our MT-DDAMFN (and this difference is insignificant~\cite{savchenko2019criterion}). However, our main objective was to obtain the models that can serve as reliable feature extractors for multiple downstream tasks. Let us demonstrate their advantages using data from the ABAW competition in the following subsections.

\subsection{VA Estimation, EXPR Recognition, and AU Detection}

\begin{table}
\scriptsize
 \centering
 \begin{tabular}{cccc}
 \toprule
Model & CCC\_V & CCC\_A& $P_{VA}$  \\
 \midrule
MT-DDAMFN & 0.412 & 0.230 & 0.321\\
MT-EmotiMobileViT & 0.403 & 0.244 & 0.324\\
MT-EmotiMobileFaceNet & \bf 0.413 & \bf 0.266 & \bf 0.339\\
MT-EmotiEffNet & 0.404 & 0.248 & 0.326\\
 \bottomrule
 \end{tabular}
 \caption{Results of pre-trained Valence-Arousal prediction models on the Aff-Wild2’s validation set. The best result is marked in bold.}
 \label{tab:va_pretrained}
\end{table}

\begin{table}
\scriptsize
 \centering
 \begin{tabular}{ccccc}
 \toprule
& \multicolumn{2}{c}{all classes}& \multicolumn{2}{c}{w/o ``Other''} \\
Model & F1-score & Accuracy & F1-score & Accuracy \\
 \midrule
EmotiEffNet-B2 & 0.229 & 0.282 & 0.320 & 0.443\\
DDAMFN & 0.244 & 0.315 & 0.362 & 0.502\\
MT-DDAMFN & 0.245 & \bf 0.340 & 0.366 & \bf 0.547\\
MT-EmotiMobileViT & 0.248 & 0.287 & 0.330 & 0.434\\
MT-EmotiMobileFaceNet & 0.250 & 0.325 & 0.354 & 0.513\\
MT-EmotiEffNet & 0.254 & 0.324 & 0.381 & 0.522\\
EmotiEffNet-B0 & \bf 0.257 & 0.325 & \bf 0.383 & 0.522\\
 \bottomrule
 \end{tabular}
 \caption{Results of pre-trained facial expression recognition models on the Aff-Wild2’s validation set. The best result is marked in bold.}
 \label{tab:expr_pretrained}
\end{table}

The results of the validation part of three conventional tasks of ABAW, namely, VA prediction, EXPR classification, and AU detection, are shown in Tables~\ref{tab:va}-~\ref{tab:au}. We compare our results with the baselines of the challenge organizers~\cite{kollias20246th} and several papers from the ABAW CVPR 2023 workshop.

Here, first, we significantly improved the metrics for VA estimation compared to our previous attempt with EfficientNet features~\cite{savchenko2023cvprw}. MT-DDAMFN achieves the top performance. It is important to emphasize that it has 2\$ greater mean CCC when compared to the initial DDAMFN, thus showing the benefits of our training procedure (Subsection~\ref{subsec:training}). It is also remarkable that the quality of the largest EmotiEffNet-B2 is the worst among our models, though it also reaches very high accuracy on AffectNet (Table~\ref{tab:affectnet}). This result highlights the need to carefully verify that the facial analysis model works in various domains and cross-dataset environments.

In contrast to the first task, facial expression recognition and AU detection do not have significant gains in using the features from the models trained in a multi-task fashion (\ref{eq:1}). Interestingly, wav2vec 2.0 embeddings work reasonably well for the EXPR challenge, leading to the top performance of simple blending~\cite{savchenko2020ad} of predictions from top visual (EmotiEffNet-B0) and audio models.

In addition, we decided to demonstrate the quality of the pre-trained models for VA prediction (Table~\ref{tab:va_pretrained}) and expression classification task (Table~\ref{tab:expr_pretrained}) without using the training set. In the latter case, the main problem is the differences in classes: AffectNet contains the Contempt category, while Aff-Wild2 includes many examples with the ``Other'' label. We studied two possibilities to map the classes: 1) assign all Contempt predictions to Other, and 2) remove the Contempt class from predictions and Other class from the validation set. Here, the best quality of VA estimation is obtained by MT-MobileFaceNet model. The highest F1-score of EXPR recognition is obtained by EmotiEffNet-B0, whose embeddings were also classified better than other models (Table~\ref{tab:expr}). However, the top accuracy for seven classes is achieved by the MT-DDAMFN model, which is 4\% higher when compared to the initial DDAMFN~\cite{zhang2023dual}.

\subsection{CE Recognition}

\begin{table}
\scriptsize
 \centering
 \begin{tabular}{c|c|ccc}
 \toprule
         & & \multicolumn{3}{c}{Clustering} \\
Model & Scores & Scores & Embeddings & Audio \\
 \midrule
DDAMFN & 0.1847 & 0.1516 & 0.2362 & 0.1792\\
EmotiEffNet-B0 & 0.1186 & 0.8419 & 1.0230 & 0.1681\\
EmotiEffNet-B0 (EXPR\_ft) & 0.5643 & \bf 0.0489 & 1.1008 & 0.2105\\
EmotiEffNet-B2 & 0.4968 & 0.1674 & 0.1791 & 0.1849\\
MT-EmotiEffNet & 0.0830 & 0.2161 & 0.5211 & 0.1681\\
MT-EmotiEffNet (EXPR\_ft) & 0.1934 & 0.1715 & 0.3206 & \bf 0.0368\\
MT-EmotiMobileFaceNet & \bf 0.0603 & 0.3316 & 0.5846 & 0.1394\\
MT-EmotiMobileViT & 0.1689 & 0.1265 & \bf 0.0975 & 0.1646\\
MT-DDAMFN & 0.2687 & 0.2037 & 0.2408 & 0.1503\\
 \bottomrule
 \end{tabular}
 \caption{The Kullback-Leibler divergence between real and predicted class probabilities for CE recognition.}
 \label{tab:ce}
\end{table}

\begin{figure}[t]
  \centering
  \includegraphics[width=0.95\linewidth]{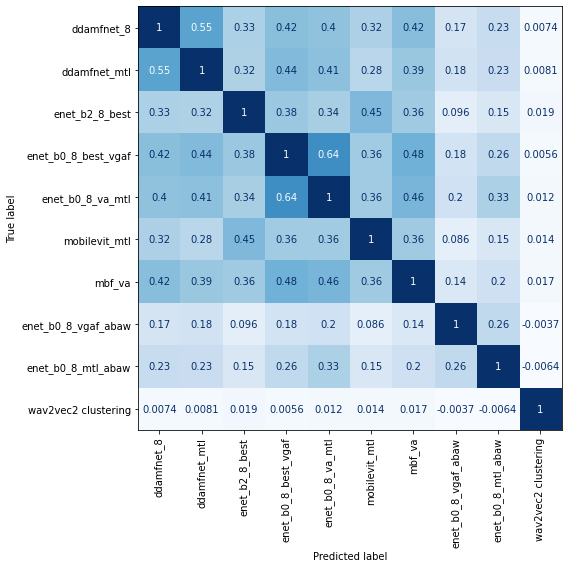}
   \caption{Kappa Cohen scores for CE predictions}
   \label{fig:ce_kappa}
\end{figure}

As we mentioned in Subsection~\ref{subsec:ce_model}, the CE prediction task has no baselines or direct metrics. Hence, we measure the indirect metric of class balance. In addition to our initial models, we used the neural networks trained to predict facial expressions using data from the EXPR challenge (hereinafter, ``(EXPR\_ft)''). We used either maximal scores at the output of our models or performed clustering of 1) scores from the final layer, 2) embeddings from the penultimate layer, and 3) wav2vec 2.0 audio embeddings.

The Kullback-Leibler (KL) divergence between actual and predicted class probabilities for all our models is shown in Table~\ref{tab:ce}. As one can notice, the KL divergence in several cases is relatively high, caused by a significant class imbalance of our predictions. It seems that MT-MobileFaceNet achieves the best balance, though the results of the MT-EmotiEffNet-B0 are also very low~\cite{savchenko2022mt}. 

To better compare our predictions, we present Cohen's kappa coefficient, which typically measures the inter-rater reliability (Fig.~\ref{fig:ce_kappa}). The most consistent with other models are EmotiEffNet-B0, MT-EmotiEffNet-B0, and MT-MobileFaceNet. Moreover, the results of clustering seem to be inconsistent with other models, so we do not expect this approach to be as accurate as other models.

\subsection{EMI Estimation}

\begin{table*}
\footnotesize
 \centering
 \begin{tabular}{cccccccp{1cm}cc}
 \toprule
 Modality & Model & Features & PCC $\overline \rho$ & Admiration & Amusement & Determination & Empathic Pain & Excitement & Joy \\
 \midrule
Faces & Baseline ViT~\cite{kollias20246th} & Embeddings & 0.09 & -& -& -& -& -& -\\
Audio & Wav2Vec2~\cite{kollias20246th} & Embeddings& 0.24& -& -& -& -& -& -\\
Audio+ & ViT+ & Embeddings & 0.25& -& -& -& -& -& -\\
Video & Wav2Vec2~\cite{kollias20246th} \\
\hline
Faces & MobileFaceNet  & Embeddings (mean) & 0.0734 & 0.0235 & 0.0542 & 0.0645 & 0.0837 & 0.1053 & 0.1093 \\
 & (VggFace2) & Embeddings (STAT) & 0.0972 & 0.0374 & 0.1008 & 0.0981 & 0.0972 & 0.1320 & 0.1175 \\ \hline

  & & Embeddings (mean) & 0.1619 & 0.0139 & 0.2515 & 0.1211 & 0.0841 & 0.2373 & 0.2641 \\
Faces  & DDAMFN & Embeddings (STAT) & 0.1603 & 0.0595 & 0.2169 & 0.1355 & 0.0687 & 0.2245 & 0.2565 \\
 & & Scores (mean) & 0.1640 & 0.0174 & 0.2462 & 0.1257 & 0.0740 & 0.2438 & 0.2770 \\
  & & Scores (STAT) & 0.1684 & 0.0354 & 0.2461 & 0.1304 & 0.0634 & 0.2426 & 0.2927 \\ \hline
 
  & & Embeddings (mean) & 0.1647 & 0.0472 & 0.2387 & 0.1272 & 0.1017 & 0.2225 & 0.2508 \\
Faces  & EmotiEffNet & Embeddings (STAT) & 0.1658 & 0.0596 & 0.2308 & 0.1318 & 0.0743 & 0.2373 & 0.2611 \\
 & -B0 & Scores (mean) & 0.1597 & 0.0163 & 0.2342 & 0.1315 & 0.0708 & 0.2281 & 0.2765 \\
  & & Scores (STAT) & 0.1645 & 0.0186 & 0.2477 & 0.1277 & 0.0787 & 0.2278 & 0.2863 \\ \hline
 
  & MT-& Embeddings (mean) & 0.1632 & 0.0162 & 0.2336 & 0.1239 & 0.1001 & 0.2339 & 0.2715 \\
Faces & EmotiEffNet & Embeddings (STAT) & 0.1673 & 0.0349 & 0.2318 & 0.1379 & 0.0877 & 0.2428 & 0.2683 \\
 & -B0 & Scores (mean) & 0.1584 & 0.0275 & 0.2115 & 0.1258 & 0.0805 & 0.2273 & 0.2776 \\
  & & Scores (STAT) & 0.1590 & 0.0188& 0.2335& 0.1150& 0.0729 & 0.2312 & 0.2828 \\ \hline
 
  & MT-& Embeddings (mean) & 0.1644 & 0.0379 & 0.2314 & 0.1387 & 0.0781 & 0.2334 & 0.2672 \\
Faces & EmotiMobile- & Embeddings (STAT) & 0.1683 & 0.0433 & 0.2459 & 0.1347 & 0.0779 & 0.2382 & 0.2699 \\
  & ViT& Scores (mean) & 0.1642 & 0.0321 & 0.2484 & 0.1490 & 0.0674 & 0.2399 & 0.2481 \\
  & & Scores (STAT) & 0.1727 & 0.0621 & 0.2548 & 0.1430 & 0.0624 & 0.2398 & 0.2738 \\ \hline

 & & Embeddings (mean) & 0.1628 & 0.0289 & 0.2385 & 0.1281 & 0.0761& 0.2363 & 0.2689 \\
Faces & MT- & Embeddings (STAT) & 0.1723 & 0.0613 & 0.2319 & 0.1282 & 0.1064 & 0.2446 & 0.2610 \\
  & DDAMFN& Scores (mean) & 0.1682 & 0.0408 & 0.2333 & 0.1387 & 0.0825 &0.2429 & 0.2710 \\
  & & Scores (STAT) & 0.1703 & 0.0289 & 0.2450 & 0.1298 & 0.0878 & 0.2410 & 0.2895 \\ \hline

  & MT-& Embeddings (mean) & 0.1518 & 0.0215 & 0.2288 & 0.1140 & 0.0692 &  0.2299 & 0.2476 \\
Faces & Emoti- & Embeddings (STAT) & 0.1646 & 0.0557 & 0.2380 & 0.1303 & 0.0703 & 0.2325 & 0.2605 \\
 & MobileFaceNet & Scores (mean) & 0.1667 & 0.0276 & 0.2367 & 0.1336 & 0.0807 & 0.2516 & 0.2699 \\
  & & Scores (STAT) & 0.1732 & 0.0285 & 0.2498 & 0.1318 & 0.097 & 0.2543 & 0.2776 \\ \hline
Audio & wav2vec 2.0  & Embeddings (mean) & 0.1514 & 0.2153 & 0.11760 & 0.1834 & 0.1426 & 0.1275 & 0.1219 \\
 & & Embeddings (STAT) & 0.2311 & 0.3006 & 0.1659 & 0.2559 & 0.3198 &0.1844 & 0.1602 \\ \hline
Audio +& &MT-DDAMFN& 0.2767 & 0.2993 & 0.3079 & 0.2230 & 0.2672 & 0.3008 & 0.2546  \\
Video & wav2vec 2.0 +  & MT-EmotiMobileViT& 0.2829 & 0.3011 & 0.2968 & 0.2595 & 0.3074 & 0.3171 & 0.2152 \\
& & MT-EmotiMobileFaceNet& \bf 0.2898 & 0.3041 & 0.3004 & 0.2584 & 0.3148 & 0.3160 & 0.2452 \\
 \bottomrule
 \end{tabular}
 \caption{EMI Estimation Pearson's correlation on the Hume-Vidmimic2's validation set.}
 \label{tab:eri}
\end{table*}

As the previous edition of EMI at the ABAW-5~\cite{kollias2023abaw2} used much more training data, our results are not directly compared with participants of that challenge. We can only compare with the audio/visual baselines obtained by ViT (Visual Transformer) and wav2vec 2.0 features. The results of our ablation experiments for the EMI task are presented in Table~\ref{tab:eri}. 

Here, our facial models are 6-8\% more accurate than the ViT baseline. However, the audio features are classified much better. We used a simpler approach for processing acoustic features, thus leading to 1\% less macro-averaged Pearson correlation $\overline \rho$. However, our best ensemble is 4\% more accurate. Like most previous experiments, multi-task learning loss (\ref{eq:1}) leads to a 0.5\% better MT-DDAMFN model. The 40-dimensional scores (logits) from the final layer of our model are recognized as not worse than high-dimensional embeddings from the penultimate layer. Finally, STAT features are typically better than the traditional average pooling of frame-level features. 

\section{Conclusion}
\label{sec:conclusion}
To conclude, we introduce several novel lightweight models trained in the multi-task framework (\ref{eq:1}) to simultaneously predict facial expression, valence, and arousal on a static photo. The weights of the neural network and the training source code to reproduce the experiments for the presented approach are publicly available\footnote{\url{https://github.com/HSE-asavchenko/face-emotion-recognition/tree/main/src/ABAW/ABAW6}}. It was experimentally demonstrated that our models reach near state-of-the-art results on conventional AffectNet benchmark (Table~\ref{tab:affectnet}). Moreover, these models extract emotional features that can be used in various downstream tasks. We demonstrated the results for all five tasks of the 6 ABAW challenge~\cite{kollias20246th}, which are essentially better when compared to baselines. For example, our best models achieved the following quality on official validation sets: CCC for VA estimation $P_{VA}=0.568$ (0.35 greater than baseline VGGFACE, Table~\ref{tab:va}), macro-averaged F1 scores for FER $P_{EXPR}=0.434$ (0.19 better than baseline VGGFace with MixAugment, Table~\ref{tab:expr}) and AU detection $P_{AU}=0.545$ (0.15 greater than baseline, Table~\ref{tab:au}). In addition, the best facial model for EMI estimation reaches macro-averaged Pearson correlation $\overline \rho=0.173$ (0.08 better than baseline ViT, Table~\ref{tab:eri}).  

It is important to emphasize that we are not required to fine-tune the model on a new dataset, so only a simple feed-forward neural network should be trained on top of our features. Though this can lead to less accurate models on concrete datasets, we firmly believe that obtaining the facial models that should analyze affective behavior in unconstrained environments for various datasets is essential.

{
    \small
    \bibliographystyle{ieeenat_fullname}
    \bibliography{main}
}

\end{document}